\documentclass{article}
\usepackage{ijcai24}

\usepackage{times}
\usepackage{soul}
\usepackage{url}
\usepackage{hyperref}
\usepackage[utf8]{inputenc}
\usepackage[small]{caption}
\usepackage{graphicx}
\usepackage{amsmath}
\usepackage{amssymb}
\usepackage[table,xcdraw]{xcolor}
\usepackage{amsthm}
\usepackage{booktabs}
\usepackage{algorithm}
\usepackage{algorithmic}
\usepackage[most]{tcolorbox}
\usepackage{multirow}
\usepackage[most]{tcolorbox}
\usepackage[switch]{lineno}

\title{Exploring Hierarchical Molecular Graph Representation in Multimodal LLMs}

\author{
    Chengxin Hu\textsuperscript{\rm 1*},
    Hao Li\textsuperscript{\rm 2*},
    Yihe Yuan\textsuperscript{\rm 1*},
    Jing Li\textsuperscript{\rm 3,4$\dagger$},
    Ivor Tsang\textsuperscript{\rm 3,4,5}
    \affiliations
    \textsuperscript{\rm 1}National University of Singapore, Singapore \\
    \textsuperscript{\rm 2}Independent Researcher \\
    \textsuperscript{\rm 3}Institute of High-Performance Computing, Agency for Science, Technology and Research, Singapore\\
    \textsuperscript{\rm 4}Centre for Frontier AI Research, Agency for Science, Technology and Research, Singapore\\
    \textsuperscript{\rm 5}College of Computing and Data Science, Nanyang Technological University, Singapore
    \emails
    e1324268@u.nus.edu,
    haolicq.ai.research@gmail.com,
    e1324267@u.nus.edu,
    Ivor.tsang@gmail.com,
    kyle.jingli@gmail.com
}

\begin{document}

\maketitle

\makeatletter
\def\thefootnote{\fnsymbol{footnote}}
\footnotetext[1]{Equal contribution, listing order is alphabetic.}
\footnotetext[2]{Corresponding author}
\def\thefootnote{\arabic{footnote}}
\setcounter{footnote}{0}
\makeatother

\begin{abstract}
    Following the milestones in large language models (LLMs) and multimodal models, we have seen a surge in applying LLMs to biochemical tasks. Leveraging graph features and molecular text representations, LLMs can tackle various tasks, such as predicting chemical reaction outcomes and describing molecular properties. However, most current work overlooks the \emph{multi-level nature} of the graph modality, even though different chemistry tasks may benefit from different feature levels. In this work, we first study the effect of feature granularity and reveal that even reducing all GNN-generated feature tokens to a single one does not significantly impact model performance. We then investigate the effect of various graph feature levels and demonstrate that both the quality of LLM-generated molecules and model performance across different tasks depend on different graph feature levels. Therefore, we conclude with two key insights: (1)~current molecular-related multimodal LLMs lack a comprehensive understanding of graph features, and (2)~static processing is not sufficient for hierarchical graph feature. We share our findings in detail, with the hope of paving the way for the community to develop more advanced multimodal LLMs for incorporating molecular graphs.
\end{abstract}
\definecolor{yello1}{RGB}{255,241,220}
\definecolor{pink1}{RGB}{242, 218, 223}
\section{Introduction}

In the 2020s, the community has shown increasing interest in leveraging the vast world knowledge of large language models (LLMs)~\cite{brown2020language,glm2024chatglm,abdin2024phi,touvron2023llama}. By utilizing powerful text tools such as SELFIES~\cite{krenn2020self} and SMILES~\cite{weininger1988smiles} to convert molecular formulas into textual strings, scientists can tap into the world knowledge of LLMs to address chemical problems.  Multimodal LLMs have now shown the ability to richly describe molecular properties and predict chemical reactions with reasonable precision. For example, GIT-MOL~\cite{liu2024git} fine-tuned state-of-the-art LLMs within multimodal frameworks, using information from various modalities to enhance the models. Following this, InstructMol~\cite{cao2023instructmol} further capitalized on instruction tuning and LoRA-based~\cite{hu2021lora} fine-tuning across different downstream tasks, creating a multimodal AI chemistry expert capable of solving a variety of chemical challenges. These achievements undeniably highlight the vast potential of Multimodal LLMs in the field of biochemistry.

However, existing works using graph features in multimodal frameworks overlook the multi-level nature of molecular graph representation. InstructMol~\cite{cao2023instructmol}, for instance, only utilizes node embeddings from its graph encoder, as does MoleculeSTM~\cite{liu2023multi}. Notice that unlike image modality, molecular graphs contain more complex structural information and semantic information related to biological and chemical contexts. We then argue that considering only a single level of information within the molecule, particularly at the atomic level, is insufficient for a model to fully comprehend the meaning of the entire molecule. For instance, the "-OH" functional group primarily contributes to a molecule’s water solubility. If we only consider atomic-level information, the model may struggle to capture the impact of the "-OH" group on downstream tasks. This limitation hinders the development of more powerful multimodal LLMs for moleculars.

This brings us to a few key questions: \textit{How do different levels of features impact various tasks? Which level is the most critical? Can combining multiple levels of features improve the performance of multimodal molecular models?}

To answer the above questions, we designed a multi-level graph encoder inspired by MoleculeSTM~\cite{liu2023multi}. Our GNN produces three levels of features—node, motif, and graph levels—in a single forward pass, using an improved virtual node approach to establish global information at the graph level. To comprehensively assess the impact of multi-level information on LLM performance, we conducted a fine-grained feature test. Specifically, we applied average pooling to the extracted features, creating three fusion methods: no reduction (no pooling), hierarchical reduction (pooling each level separately), and all reduction (pooling all features into a single token). We found that even pooling all feature tokens into one does not significantly harm LLM performance. Next, we extracted each of the three levels produced by the GNN and fed them individually into the LLM during training. We observed that some levels improved exact match score, achieving a perfect match with the ground truth molecules, while others enhanced overall similarity between generated and target molecules. From these experiments we also see that the optimal feature level essentially varied across tasks. Based on these findings, we provide our insights to this challenge which points out the possible solutions for future studies. We note a parallel work~\cite{chen2024hight} revealed that hierarchical graph features help mitigate hallucinations in LLMs in molecular-related experiments, while our work focuses on a comprehensive investigation of the effects of different levels of graph features in multimodal LLMs. Our contributions are summarized as follows. 
\begin{itemize}
    \item We pioneer a comprehensive study on hierarchical graph representations for molecules in the context of training multimodal LLMs. We highlight its practical significance through revisiting LLMs-for-Molecular research.
    \item We propose a Multi-level Molecular Multimodal LLM (M$^3$LLM) architecture, in which we first perform representation extraction for graph modality and then learn to project for the alignment with LLMs.
    \item We evaluate the impact of each feature level across five key downstream tasks with various feature fusion strategies (no reduction, hierarchical reduction, and all reduction) being applied, and  provide key insights by analyzing the experimental results. 
\end{itemize}

\section{Preliminaries}
\subsection{Tasks statement}
To assess LLMs' grasp of molecular graphs, we consider three categories of molecular-related tasks: Molecular Reaction Prediction, Molecular Property Prediction, and Molecular Description Generation. In each, by leveraging SMILES and SELFIES-—two one-dimensional textual representations of molecules, we extract detailed information from both graph and text modalities for LLM processing, yielding task-specific outputs. For Molecular Reaction Prediction, we explore forward reaction prediction, and reagent prediction, inferring one component from the other two among reactants, reagents, and products. Data is drawn from~\cite{USPTO2020}—a broadly used, extensive reaction dataset refined from U.S. patents with wide bio-pharmaceutical applications. For Molecular Caption Generation (Molcap), rather than following~\cite{fang2023mol}, we use ChEBI-20~\cite{edwards2021text2mol}—optimized via PubChem~\cite{kim2023pubchem}—to deliver comprehensive molecular descriptions.

\subsection{Molecular Multimodal LLMs}
Regarding molecular multimodal LLMs (MLLMs) from research, there is a trend toward unifying the architectures within the Blip-2~\cite{li2023blip} and LLaVA~\cite{liu2024visual} frameworks, effectively standardizing them as an encoder-projector-LLM system. Typically, the encoder processes non-text modalities, and the projector conducts alignment before feeding them to the LLM for downstream tasks. \cite{liu2023molca} first proposed a Blip-2-based molecular MLLM, and \cite{li2024towards} subsequently introduced LLM tasks in 3D molecular scenarios. \cite{cao2023instructmol} pioneered a LLaVA-based molecular MLLM, while a contemporary work~\cite{chen2024hight} incorporated molecular node and motif-level information for domain-specific tasks. Recent research~\cite{hu2025omni} also presented a scalable, unified LLM framework for direct instruction tuning using a homogeneous molecular MLLM architecture. However, unlike Vision Large Language Models (VLLMs), molecular MLLMs include molecular text as auxiliary data, meaning they do not rely solely on molecular graphs. So far, it remains to be determined whether LLMs truly grasp molecular graphs and to what extent this understanding occurs.

\section{Methods}
\begin{figure*}[t]
    \centering
    \includegraphics[scale=1]{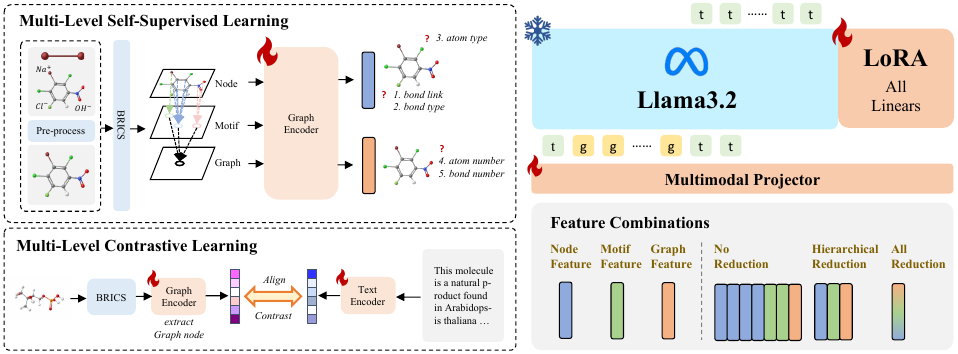}
    \caption{Overview of M$^3$LLM Architecture. \textbf{Left:} A pipeline comprising GNN-based hierarchical graph dynamic segmentation, multi-level self-supervised learning, and multi-level contrastive learning. \textbf{Right:} A two-stage training pipeline of M$^3$LLM.}
    \label{fig:mainfig}
\end{figure*}
\subsection{Overview}
To facilitate the investigation of how large language models interpret molecular graph features—and following~\cite{cao2023instructmol}—we have developed \textbf{M$^\textbf{3}$LLM}, a LLaVA-inspired, multi-level molecular multimodal large language model. Let \(X_S\) and \(X_I\) denote the SELFIES representation of the molecule and the task instruction, respectively, and let \(\mathcal{G}\) represent the graph input data. Our model is formulated as :
\begin{equation}
    R = \text{LM}_{\theta}(X_S, X_I, f_p(f_g(\mathcal{G})))
\end{equation}
where $R$ is the reponse of the model, $\text{LM}$ is a pre-trained large language model parameterized by $\theta$, $f_g$ is the graph encoder and $f_p$ is a projector used to align graph feature with the LLM. 

We conduct two stages to train M$^3$LLM (Refer to the right panel of Figure~\ref{fig:mainfig}):

\noindent\textbf{Alignment Pre-training}. 
In this stage, we freeze all parameters except for the projector, allowing the projector to learn how to align the graph feature space with the embedding space of the large language model. During this phase, we use the PubChem~\cite{kim2023pubchem} dataset to construct a molecular description task, aligning graph and text features on over 300,000 samples.

\noindent\textbf{LoRA Fine-tuning}
After alignment, we use LoRA~\cite{hu2021lora} for fine-tuning, freezing the weights of the graph encoder and the large language model while activating the weights of the projector and LoRA. This process enables the large language model to handle multimodal data. We fine-tune on five different tasks, resulting in five distinct LoRA adapters.

\subsection{Designing Hierarchical Graph Encoder}
The information inherent in graph-based molecular data is more complex compared to other modalities. Most existing message-passing-based graph encoders focus on single-level information, such as node-level details. While higher-dimensional information extraction can be achieved through multi-layer models, we believe this approach still lacks fine-grained molecular semantics and structural information, which can pose challenges when aligning with textual data. This makes it difficult for subsequent LLMs to truly understand the fine-grained meaning of input molecular graphs. Additionally, existing molecular datasets lack a refined summary of molecular low-dimensional structure and semantic information. To address this, we designed a dynamic molecular graph fine-grained segmentation algorithm, which automatically segments molecular graphs into different levels. We also redesigned the training pipeline to achieve finer-grained alignment between molecules and text, while ensuring the performance of GNNs on molecule-centric tasks. Our GNN design can be referred to the left panel of Figure~\ref{fig:mainfig}. 
\subsubsection{Hierarchical Graph Dynamic Segmentation}
To obtain the hierarchical features $\mathcal{G}$ of the molecular graph \( G \), we first directly obtain the node embeddings \( V \) of the molecule. Then, we apply the Breaking of Retrosynthetically Interesting Chemical Substructures (BRICS) algorithm to extract the motif information \( M \) from the molecule. Without loss of generality, we treat the extracted motifs as \texttt{"}nodes\texttt{"} \( V_m \) (motifs are unified as special nodes and tokenized identically to regular nodes) and add them to the molecular graph, creating node-motif edges \( E_m \) within their respective coverage. For graph-level, unlike virtual nodes, we introduce a virtual graph node \( V_g \) into the molecular graph, while only considering the connection between the graph node \( V_g \) and the motif nodes via edges \( E_g \). The final multi-level molecular feature representation is denoted as \( G = (V, E) \), where \( V = \{V, V_m, V_g\} \) and \( E = \{E, E_m, E_g\} \).

\subsubsection{Multi-level Self-supervised Learning}
To enable the GNN to efficiently learn the complex inner information of molecular graphs, we standardized and purified the SMILES of molecules obtained from PubChem~\cite{kim2023pubchem}, eliminated free ions, and discarded molecules with fewer than five atoms. As a result, we acquired 1 million high-quality SMILES. For the refined PubChem SMILES data, we leverage the Hierarchical Graph Dynamic Segmentation algorithm to extract molecular information at the atom, motif, and graph levels. At the atomic level, we introduce three generative pre-training tasks that exploit the molecule's atomic and bond characteristics, each with its own dedicated prediction head. We then optimize using three distinct cross-entropy losses:

\begin{equation}
L_{\text{link}} = -\sum_{v_i,v_j\in V}\Bigl[y_{ij}\log\hat{y}_{ij} + (1-y_{ij})\log(1-\hat{y}_{ij})\Bigr]
\end{equation}

\begin{equation}
L_{\text{AT}} = -\frac{1}{|V|}\sum_{v\in V}\sum_{a=1}^{A_{\text{atom}}}y_{v,a}\log\hat{y}_{v,a}
\end{equation}

\begin{equation}
L_{\text{BT}} = -\frac{1}{|E|}\sum_{e\in E}\sum_{l=1}^{L_{\text{bond}}}y_{e,l}\log\hat{y}_{e,l}
\end{equation}
where \(y_{ij}=1\) indicates that nodes \(i\) and \(j\) are connected, \(y_{v,a}=1\) means node \(v\) belongs to type \(a\), and \(y_{e,l}=1\) signifies that bond \(e\) is of type \(l\). The term \(\hat{y}\) denotes the corresponding predictions.

At the graph level, we formulate two predictive tasks, each accompanied by its own downstream head. The corresponding loss functions are defined as:

\begin{equation}\label{eq5}
L_{\text{AN}} =
\begin{cases}
0.5\cdot (y_{an}-\hat{y}_{an})^2, & \text{if } |y_{an}-\hat{y}_{an}| < 1 \\
|y_{an}-\hat{y}_{an}|-0.5, & \text{otherwise}
\end{cases}
\end{equation}

\begin{equation}\label{eq6}
L_{\text{BN}} =
\begin{cases}
0.5\cdot (y_{bn}-\hat{y}_{bn})^2, & \text{if } |y_{bn}-\hat{y}_{bn}| < 1 \\
|y_{bn}-\hat{y}_{bn}|-0.5, & \text{otherwise}
\end{cases}
\end{equation}
where \(y_{an}\) and \(y_{bn}\) represent the number of atoms and bonds, respectively, while \(\hat{y}\) refers to the corresponding predicted values.

\subsubsection{Multi-level Contrastive Learning}
Incorporating text attributes into the GNN has been demonstrated to be extremely advantageous for LLMs in molecular understanding and downstream tasks~\cite{tong2024cambrian,xie2024could}. We utilize over 310K molecule-text pairs from PubChem, following the approach in MoleculeSTM~\cite{liu2023multi}, as input for contrastive learning. The textual descriptions are processed via SciBERT~\cite{beltagy2019scibert} and then pooled to obtain the feature vector \( x_t \). For the molecular graph, since graph-level information becomes more granular through its interaction with atomic-level data during GNN propagation, we employ the graph node features, denoted by \( x_m \), to represent the molecule's multi-level characteristics and align them with \( x_t \) for joint training of the SciBERT and GNN models. The training objective is written as:
\begin{align}
L_{\text{MTM}} = & -\frac{1}{2}\Biggl[
\mathbb{E}_{x_m,x_t}\bigl[\log\sigma\bigl(E(x_m,x_t)\bigr)\bigr] \notag \\
& + \mathbb{E}_{x_m,x'_t}\bigl[\log\bigl(1-\sigma\bigl(E(x_m,x'_t)\bigr)\bigr)\bigr] \notag \\
& + \mathbb{E}_{x'_m,x_t}\bigl[\log\bigl(1-\sigma\bigl(E(x'_m,x_t)\bigr)\bigr)\bigr]
\Biggr],
\end{align}
where \( x_t \) and \( x_m \) represent the positive samples for the molecule-text pairs, while \( x'_t \) and \( x'_m \) are negative samples drawn from the noise distribution, and \( E(\cdot) \) denotes the inner product between vectors.

\subsection{Exploring Multi-Level Features}
\label{sec:emf}
\begin{table*}[t]
\centering
\scriptsize
\vspace{-0.01cm}
\renewcommand{\arraystretch}{1.1}
\setlength{\tabcolsep}{4.6mm}{
\begin{tabular}{lccccccc}
\toprule[1.1pt]
Method & \#Param & BLEU-2 & BLEU-4 & ROUGE-1 & ROUGE-2 & ROUGE-L & METEOR \\ \hline
\multicolumn{8}{l}{\cellcolor[HTML]{E6E7FF}Molecule Caption} \\
InstructMol-G & 6.7B & 0.466 & 0.365 & 0.547 & 0.365 & 0.479 & 0.491 \\
InstructMol-GS & 6.7B & \textbf{0.475} & \textbf{0.371} & \textbf{0.566} & \textbf{0.394} & \textbf{0.502} & \textbf{0.509} \\
\rowcolor[HTML]{EFEFEF} 
\textbf{M$^3$LLM} & \textbf{1.2B} & 0.403 & 0.306 & 0.518 & 0.342 & 0.455 & 0.450 \\ \toprule[1.1pt]
\end{tabular}
}
\setlength{\tabcolsep}{4.3mm}{
\begin{tabular}{lcccccccc}
\toprule[1.1pt]
Method & \#Param & BLEU & Exact & Levenshtein$\downarrow$ & Validity & MACCS & RDK & Morgan \\ \hline
\rowcolor[HTML]{E6E7FF} 
\multicolumn{9}{l}{\cellcolor[HTML]{E6E7FF}Forward Reaction Prediction} \\
LLaMA2 & 6.7B & 0.804 & 0.012 & 29.947 & 1.000 & 0.649  & 0.499 & 0.407 \\
Vicuna$^1$ & 13B & 0.057& 0.000 & 41.690 & 0.059 & 0.016 & 0.007 &0.006 \\
Mol-Instruction & 6.7B & 0.654 & 0.045 & 27.262 & 1.000 & 0.509 & 0.313 & 0.262 \\
HIGHT & 6.7B & 0.935 & 0.293 & 16.687 & 1.000 & 0.618 & 0.774 & 0.566 \\
InstructMol-G & 6.7B & 0.906 & 0.153 & 20.155 & 1.000 & 0.691 & 0.469 & 0.426 \\
InstructMol-GS & 6.7B & 0.967 & 0.536 & 10.851 & 1.000 & 0.878 & 0.776 & 0.741 \\
\rowcolor[HTML]{EFEFEF} 
\textbf{M$^3$LLM} & \textbf{1.2B} & \textbf{0.967} & \textbf{0.556} & \textbf{10.223} & 0.999 & \textbf{0.882} & \textbf{0.784} & \textbf{0.755} \\ \hline
\rowcolor[HTML]{E6E7FF} 
\multicolumn{9}{l}{\cellcolor[HTML]{E6E7FF}Reagent Prediction} \\
Mol-Instruction  & 6.7B & 0.224 & 0.044 & 23.167 & 1.000 & 0.364 & 0.237 & 0.213 \\
HIGHT & 6.7B & 0.482 & 0.067 & 27.167 & 1.000 & 0.346 & 0.462 & 0.303 \\
InstructMol-G & 6.7B & \textbf{0.890} & 0.070 & 24.732 & 1.000 & \textbf{0.691} & \textbf{0.469} & \textbf{0.426} \\
InstructMol-GS & 6.7B & 0.610 & 0.129 & 19.664 & 1.000 & 0.539 & 0.444 & 0.400 \\
\rowcolor[HTML]{EFEFEF} 
\textbf{M$^3$LLM} & \textbf{1.2B} & 0.646 & \textbf{0.131} & \textbf{18.720} & 1.000 & 0.528 & 0.426 & 0.388 \\ \toprule[1.1pt]
\end{tabular}
}
\caption{Results on two downstream tasks. $^1$: 5-shot In-Context Learning results from InstructMol. All \#Param shows the number of parameters of the LLM backbones.}
\label{tab:reproduction}
\end{table*}

To explore the impact of different feature levels, we first established our baseline. We selected the 1B-sized LLaMA-3.2~\cite{dubey2024llama} as the language model backbone and re-evaluated the performance of the InstructMol model on five tasks. The results are presented in Table~\ref{tab:reproduction}.
    
Experimental results show that M$^3$LLM achieves promising performance compared to InstructMol and other baselines while maintaining a small model size. The model with the smaller backbone showed a significant increase in performance on forward prediction and reagent prediction tasks. In the retrosynthesis task, molecular similarity to the ground truth decreased slightly, while performance dropped more noticeably in the molecule caption task. However, prediction accuracy in the property prediction task saw a modest increase.

\textit{Remark: Notably, the smaller model generated some invalid molecules. In the retrosynthesis task, the model produced four incorrect molecules out of 1,000 test samples, suggesting potential issues with the smaller model’s understanding of molecules and their SELFIES representations. The significant performance drop in the Molcap task reflects the smaller model’s limitations in language capability.}

In the following two sections, we conduct experiments on three representative tasks: \textbf{forward reaction prediction, reagent prediction, and molcap}.
\subsubsection{Exploring the Impact of Token Reduction}
\begin{table*}[t]
\centering
\scriptsize
\vspace{-0.01cm}
\renewcommand{\arraystretch}{1.1}
\setlength{\tabcolsep}{1.5mm}{
\begin{tabular}{l|lccccccc}
\hline
Task & Method & BLEU & Exact & Levenshtein & Validity & MACCS & RDK & Morgan \\ \hline
 & \cellcolor[HTML]{E4FFF4}No reduction & \cellcolor[HTML]{E4FFF4}0.967 & \cellcolor[HTML]{E4FFF4}0.556 & \cellcolor[HTML]{E4FFF4}\textbf{10.223} & \cellcolor[HTML]{E4FFF4}0.999 & \cellcolor[HTML]{E4FFF4}\textbf{0.882} & \cellcolor[HTML]{E4FFF4}\textbf{0.784} & \cellcolor[HTML]{E4FFF4}\textbf{0.755} \\
 & Hierarchical reduction & \textbf{0.970(+0.003)} & 0.552(-0.004) & 10.472(+0.249) & \textbf{1.000(+0.001)} & 0.879(-0.003) & 0.778(-0.006) & 0.750(-0.005) \\
\multirow{-3}{*}{\begin{tabular}[c]{@{}l@{}}Forward\\ Reaction\\ Prediction\end{tabular}} & All reduction & 0.966(-0.001) & 0.556(+0.000) & 10.968(+0.745) & 0.999(+0.000) & 0.876(-0.006) & 0.770(-0.014) & 0.742(-0.013) \\ \hline
 & \cellcolor[HTML]{E4FFF4}No reduction & \cellcolor[HTML]{E4FFF4}0.646 & \cellcolor[HTML]{E4FFF4}0.131 & \cellcolor[HTML]{E4FFF4}18.720 & \cellcolor[HTML]{E4FFF4}\textbf{1.000} & \cellcolor[HTML]{E4FFF4}0.528 & \cellcolor[HTML]{E4FFF4}0.426 & \cellcolor[HTML]{E4FFF4}0.388 \\
 & Hierarchical reduction & 0.627(-0.019) & 0.120(-0.011) & 19.446(+0.726) & 1.000(+0.000) & 0.518(-0.010) & 0.418(-0.008) & 0.378(-0.010) \\
\multirow{-3}{*}{\begin{tabular}[c]{@{}l@{}}Reagent\\ Prediction\end{tabular}} & All reduction & \textbf{0.668(+0.022)} & 0.127(-0.004) & \textbf{17.875(-0.845)} & 1.000(+0.000) & 0.534(+0.006) & \textbf{0.432(+0.006)} & \textbf{0.397(+0.009)} \\ \hline
\end{tabular}
}
\setlength{\tabcolsep}{3.0mm}{
\begin{tabular}{l|lcccccc}
\hline
Task & Method & BLEU-2 & BLEU-4 & ROUGE-1 & ROUGE-2 & ROUGE-L & METEOR \\ \hline
 & \cellcolor[HTML]{E4FFF4}No Reduction & \cellcolor[HTML]{E4FFF4}\textbf{0.403} & \cellcolor[HTML]{E4FFF4}\textbf{0.306} & \cellcolor[HTML]{E4FFF4}\textbf{0.518} & \cellcolor[HTML]{E4FFF4}\textbf{0.342} & \cellcolor[HTML]{E4FFF4}\textbf{0.455} & \cellcolor[HTML]{E4FFF4}\textbf{0.450} \\
 & Hierarchical Reduction & 0.400(-0.003) & 0.302(-0.004) & 0.517(-0.001) & 0.341(-0.001) & 0.454(-0.001) & 0.448(-0.002) \\
\multirow{-3}{*}{Molcap} & All Reduction & 0.401(-0.002) & 0.304(-0.002) & 0.515(-0.003) & 0.340(-0.002) & 0.453(-0.002) & 0.446(-0.004) \\ \hline
\end{tabular}
}
\caption{Impact of different reduction methods, The bold number is the best metric among three reduction methods, we also calculated the difference between the method and baseline.}
\label{tab:reduction}
\end{table*}
When using multi-level features, a natural question arises: how should we appropriately handle the feature tokens from the three levels? Should we directly feed all tokens into the LLM? Can we summarize features from the three levels? Or even merge all features into a single token?

Here, we test three feature fusion methods, which we also refer to as token reduction. Let $h_\mathcal{G} = \{\{n_i\}_{i=1}^{a}, \{m_j\}_{j=1}^{b}, g\}$ represents the set of features generated by the multi-level GNN, where $n_i, m_j, g\in \mathbb{R}^{1\times d_{GNN}}$, $d_{GNN}$ is the hidden dimension of the GNN, and $a$, $b$ represents the number of feature tokens in node level and motif level respectively, graph level only contains a final summarized token produced by our GNN. The projector will map the dimension of the features in $h_\mathcal{G}$ to the dimension of LLM, $n_i, m_j, g\in \mathbb{R}^{1\times d_{GNN}} \overset{\text{projector}}{\rightarrow}\mathbb{R}^{1\times d_{LLM}}$.

\paragraph{No reduction.} We concatenate all features and feed them into LLM directly
\begin{equation}
    h_{\mathbf{G}}\in \mathbb{R}^{(a+b+1)\times d_{LLM}} = f_p(\text{Concat}(h_\mathcal{G}))
\end{equation}
where $\text{Concat}$ is the concatenation operation on the first dimension.

\paragraph{Hierarchical reduction.}
We attempt to fuse feature tokens from the same feature level, compressing information from each level into a single token. Specifically, we first project the feature to $d_{LLM}$
\begin{equation}
    h_{\mathbf{n}}^{(i)} = f_p(n_i), h_{\mathbf{m}}^{(i)} = f_p(m_i), h_{\mathbf{g}} = f_p(g)
\end{equation}
where $h_{\mathbf{n}}^{(i)}, h_{\mathbf{m}}^{(i)}, h_{\mathbf{g}}\in \mathbb{R}^{1\times d_{LLM}}$ are projected features.

We then apply average pooling
\begin{equation}
    \bar h_{\mathbf{n}} = \frac{1}{a}\sum_{i=1}^a h_{\mathbf{n}}^{(i)}\quad \bar h_{\mathbf{m}} = \frac{1}{b}\sum_{i=1}^b h_{\mathbf{m}}^{(i)}
\end{equation}
finally, we concatenate the pooled results.
\begin{equation}
    h_{\mathbf{G}} \in \mathbb{R}^{3\times d_{LLM}} = \text{Concat}(\bar h_{\mathbf{n}}, \bar h_{\mathbf{m}}, h_{\mathbf{g}})
\end{equation}

\paragraph{All reduction.} Finally, to complete the experiment, we fuse all the features into one token, namely all reduction. We apply Global Average Pooling(GAP) to the features.
\begin{equation}
    h_{\mathbf{G}} \in \mathbb{R}^{1\times d_{LLM}} = \frac{1}{a+b+1}\sum_{i=1}^{a+b+1}f_p\left(h_{\mathcal{G}}^{(i)}\right)
\end{equation}

The results are shown in Table~\ref{tab:reduction}, For the forward reaction prediction task, the no-reduction method performs best, with only minor fluctuations in BLEU and Validity, and all other metrics outperforming the other two reduction methods. This may be because forward reaction prediction requires high-granularity interactions between atoms and functional groups to predict atomic additions and removals in molecules. However, in the reagent prediction task, the best method is all-reduction; except for a slightly lower Exact Match score compared to no-reduction, all other metrics favor all-reduction. This suggests that reagent prediction does not require low-level atomic information, and with all-reduction, we observe significant improvements in various metrics. For molcap, all metrics consistently perform better with no-reduction, indicating that preserving all details is essential for molecular description tasks. However, hierarchical reduction yields generally poor performance for all tasks.

\textit{Remark: Intuitively, average pooling captures global features but may lose substantial information. However, as shown in Table~\ref{tab:reduction}, the performance gap between all-reduction and no-reduction is not particularly large; even when a certain reduction method performs better, the difference is still not significant. This suggests that complete, high-granularity graph features do not necessarily lead to a more comprehensive molecular understanding by the LLM. Instead, the LLM may rely more heavily on SELFIES for reasoning, indicating that the impact of the graph modality may be less significant than that of the textual modality.}

\subsubsection{Exploring the Impact of Different Levels}
\begin{table*}[t]
\centering
\scriptsize
\vspace{-0.01cm}
\renewcommand{\arraystretch}{1.1}
\setlength{\tabcolsep}{2.2mm}{
\begin{tabular}{l|lccccccc}
\bottomrule[1.1pt]
Task & Method & BLEU & Exact & Levenshtein$\downarrow$ & Validity & MACCS & RDK & Morgan \\ \hline
\multirow{3}{*}{\begin{tabular}[c]{@{}l@{}}Forward\\ Reaction\\ Prediction\end{tabular}} 
 & Node level & \textbf{0.968(-0.001)} & 0.555(-0.002) & 10.567(+0.196) & \textbf{0.999(+0.001)} & 0.875(-0.006) & 0.773(-0.006) & 0.747(-0.003) \\
 & Motif level & \textbf{0.968(-0.001)} & 0.553(-0.004) & \textbf{10.305(-0.066)} & \textbf{0.999(+0.001)} & 0.878(-0.003) & 0.779(+0.000) & 0.750(+0.000) \\
 & Graph level & \textbf{0.968(-0.001)} & \textbf{0.565(+0.008)} & 10.442(+0.071) & \textbf{0.999(+0.001)} & \textbf{0.881(+0.000)} & \textbf{0.782(+0.003)} & \textbf{0.752(+0.002)} \\ \hline
\multirow{3}{*}{\begin{tabular}[c]{@{}l@{}}Reagent\\ Prediction\end{tabular}} 
 & Node level & 0.630(-0.025) & 0.111(-0.053) & 19.462(+1.562) & \textbf{1.000(+0.000)} & \textbf{0.617(+0.065)} & 0.405(-0.055) & 0.372(-0.049) \\
 & Motif level & \textbf{0.631(-0.024)} & 0.118(-0.046) & \textbf{19.402(+1.502)} & 0.998(-0.002) & 0.525(-0.027) & \textbf{0.427(-0.033)} & \textbf{0.385(-0.036)} \\
 & Graph level & 0.625(-0.030) & \textbf{0.136(-0.028)} & 19.657(+1.757) & \textbf{1.000(+0.000)} & 0.526(-0.026) & 0.420(-0.040) & 0.384(-0.037) \\ \toprule[1.1pt]
\end{tabular}
}
\setlength{\tabcolsep}{3.7mm}{
\begin{tabular}{l|lcccccc}
\bottomrule[1.1pt]
Task & Method & BLEU-2 & BLEU-4 & ROUGE-1 & ROUGE-2 & ROUGE-L & METEOR \\ \hline
\multirow{3}{*}{Molcap} 
 & Node level & 0.393(-0.082) & 0.294(-0.077) & 0.508(-0.058) & 0.332(-0.062) & 0.446(-0.056) & 0.438(-0.071) \\
 & Motif level & \textbf{0.395(-0.080)} & \textbf{0.298(-0.073)} & \textbf{0.514(-0.052)} & \textbf{0.336(-0.058)} & \textbf{0.452(-0.050)} & \textbf{0.444(-0.065)} \\
 & Graph level & 0.392(-0.083) & 0.295(-0.076) & 0.513(-0.053) & \textbf{0.336(-0.058)} & 0.451(-0.051) & 0.440(-0.069) \\ \toprule[1.1pt]
\end{tabular}
}
\caption{Impact of different feature levels, the bold number is the best metric among three levels, we also calculated the difference between the method and baseline.}
\label{tab:levels}
\end{table*}

Here, we further explore the impact of different feature levels on various tasks. By default, we apply all-reduction to all feature tokens, forcing the node and motif levels to be compressed into a single token, just like the graph level. We extract features from the node, motif, and graph levels from the GNN and process each separately. During training, only one level of features is fed into the LLM at a time. The results are shown in Table~\ref{tab:levels}.

In the forward reaction prediction task, we observed a significant performance boost when using graph-level features. The Exact match score improved by 0.01 compared to the second-best, node-level features, and aside from minor variations in Levenshtein, all other metrics outperformed the other two levels. In the Reagent Prediction task, the motif level showed the best performance on molecular similarity metrics, but the graph level achieved the highest Exact match, with a 0.018 improvement over the next best, motif level. For the molcap task, the motif level yielded the best results, suggesting that functional group information plays a crucial role in molecular description for this task. Comparing Tables~\ref{tab:reduction} and~\ref{tab:levels}, using only the graph level outperformed multi-level features in the forward reaction prediction and reagent prediction tasks, whereas multi-level features provided superior performance in the molcap task.

\textit{Remark: By constructing a specialized virtual node, the multi-level GNN can better summarize global information, which, in some tasks, can even outperform methods that provide fine-grained details. For example, in the forward reaction prediction task, the graph-level approach surpasses the performance of the multi-level features with the no reduction method. This indicates that global information plays a crucial role in enabling the LLM to understand molecular structures and demonstrates the effectiveness of our virtual node approach.}

\section{Experiment}
\subsection{Setup}
\paragraph{Enviroment.} All experiments were conducted using the PyTorch framework, with the Huggingface Transformers~\cite{wolf-etal-2020-transformers} and PEFT~\cite{peft} libraries, and optimized for multi-GPU parallelism with DeepSpeed~\cite{rajbhandari2020zero}. For molecule related processing, we use RDKit. We utilized four Nvidia RTX 3090 GPUs for all experiments. To ensure consistency across the experiments, we fixed the random seed to 0.

\paragraph{Implementation.} The M$^3$LLM follows the implementation of InstructMol, where graph tokens are fed into LLM without compression, i.e., with no reduction. We concatenate graph features to the beginning of the user prompt.

\paragraph{Training Details.} For three reaction tasks and property prediction task, we use a learning rate of 8e-5, we fix the batch size to 64, and trained the model for 10 epochs. For molecule caption task, we trained 6 epochs since we observed overfit when trained for more epochs, we keep the learning rate of 8e-5 and the batch size is set to 16. We use flash attention 2~\cite{dao2023flashattention2}to accelerate the training. For optimizer, we use AdamW~\cite{loshchilov2017decoupled} by default. 

\paragraph{Baselines.}
We primarily used InstructMol~\cite{cao2023instructmol} as the baseline, with MoleculeSTM~\cite{liu2023multi}, employed in InstructMol, serving as the GNN baseline. For results already reported in the literature, we directly adopted them. However, as mentioned earlier, we re-evaluated a version with a different LLM backbone to establish a new baseline.

\paragraph{Datasets.}
For the reaction prediction and property prediction tasks, we use the instruction-tuning dataset from Mol-Instructions~\cite{fang2023mol}. For the molcap task, we utilize the CHEBI-20~\cite{edwards2021text2mol} dataset. The experiments are conducted using the train-test splits as defined in the original datasets.

\subsection{Results}
For the baselines, we report in Table~\ref{tab:reproduction} the performance on five tasks. For our exploration of multi-level features, we selected three representative tasks: forward reaction prediction, reagent prediction, and molecule captioning (molcap). Table~\ref{tab:reduction} shows the impact of different reduction methods on model performance, while Table~\ref{tab:levels} presents results obtained by consistently using the all reduction method across experiments. This approach ensures that all feature levels are reduced to a single token, keeping the token count and format consistent with that of the graph level.

\subsection{Analysis}
In Section~\ref{sec:emf}, we conducted a preliminary analysis of our experimental results. In this part, we provide two key insights: \colorbox{yello1}{(1)}~LLMs lack a comprehensive understanding of graph features and are not fully capable of leveraging them effectively. \colorbox{pink1}{(2)}~Different tasks may require features from different levels, indicating a need for a dynamic projector capable of processing information across multiple levels.

\paragraph{\colorbox{yello1}{Insight 1}: LLM lacks a comprehensive understanding of graph features.}
Graph features can enhance an LLM’s understanding of molecular structures, making molecular MLLM that integrate graph features a growing area of research. Theoretically, node embeddings capture atom-level information, where a node-level token may represent an atom in a molecule and its connections to other atoms. A motif-level token, on the other hand, may represent a functional group within the molecule and its interactions with other regions. These fine-grained features provide the LLM with atom- and functional group-level insights, allowing it, for instance, to leverage atomic connectivity in tasks like forward reaction prediction, where it can understand reactants and adjust chemical bonds to predict products.

However, averaging the tokens from all three levels into a single pooled token did not result in severe performance degradation. This pooled token retains global information but lacks atom- or motif-level granularity. Supplying the LLM with this overly compressed feature set still yielded acceptable performance across three downstream tasks. This may imply that the LLM does not heavily rely on multi-level graph features or may not fully comprehend them. Instead, it likely maps inputs to outputs by leveraging relationships encoded in SELFIES representations.

This suggests that new alignment training approaches may be necessary: first, to enable the LLM to understand relationships between graph features and atomic or functional group-level structures within molecules, and second, to foster an understanding of the connections between graph features and SELFIES representations. Currently, many approaches rely on aligning graph features with molecular descriptions from natural language-trained LLMs. Such alignment may be insufficient for deep comprehension of graph features. Aligning multi-level information, and graph features and SELFIES representations, could enhance the LLM’s grasp of multimodal data.

\paragraph{\colorbox{pink1}{Insight 2}: We need to dynamically process features from different levels.}
The results from Table~\ref{tab:levels} demonstrate that, in reaction prediction tasks, graph-level features significantly enhance exact match performance, while features from other levels improve the similarity between the predicted and target molecules. For the molcap task, motif-level features yield the best performance, suggesting that both the quality of generated molecules and task performance across various tasks depend on features from different hierarchical levels. Each level—whether node, motif, or graph—contributes unique semantic information that can benefit specific tasks, making it essential to selectively leverage these layers.

However, the current linear projector only performs a straightforward projection of graph features from the GNN embedding space to the LLM embedding space, adopting a static approach that lacks adaptability. This method does not account for the semantic differences across feature levels, limiting the LLM’s ability to fully utilize each layer’s unique information. This static approach falls short in tasks requiring a nuanced balance of feature fusion, suppression, or emphasis. For example, some tasks may benefit from a stronger focus on graph-level features to capture the global structure, while others may require finer details from node- or motif-level features. A dynamic projector would enable selective fusion and adjustment of features based on the specific needs of each task and molecular structure. Such a design could enhance the LLM’s ability to extract critical task-specific information from each hierarchical level, ultimately leading to more accurate predictions and a more comprehensive understanding of molecular representations. 

We believe that implementing a dynamic projector could not only solve these challenges but also maximize the utility of each feature level, providing the LLM with richer, semantically consistent information that aligns closely with the requirements of various biochemical tasks.

\section{Conclusion}
This paper proposed a Multi-level Molecular Multimodal LLM (M$^3$LLM) architecture, which utilizes the multi-level features generated by the GNN to examine their impact on the LLM’s understanding and generation of molecular information. By studying various feature fusion methods and analyzing the performance of different feature levels across different tasks, we derive two key insights. First, current LLMs lack a comprehensive understanding of graph features, highlighting the need for improved alignment pretraining strategies to help the LLM align multi-level information and better understand the relationships between SELFIES and graph modalities. Second, features from different levels play distinct roles; thus, a projector is needed to dynamically process these features and maximize the contribution of each level.

\appendix

\section*{Ethical Statement}

There are no ethical issues.


\bibliographystyle{named}
\bibliography{ijcai24}

\section*{Appendix}
\section{More Training Details}
\label{app:traindetai}
\begin{table}[t]
\centering
\scriptsize
\vspace{-0.01cm}
\renewcommand{\arraystretch}{1.1}
\setlength{\tabcolsep}{2mm}{
\begin{tabular}{l|cccccc}
\bottomrule[1.1pt]
 & \multicolumn{1}{c|}{FWD} & \multicolumn{1}{c|}{REGT} & \multicolumn{1}{c|}{RTSTS} & \multicolumn{1}{c|}{PRPT} & \multicolumn{1}{c|}{MCP} & Pre. \\ \hline
learning rate & \multicolumn{5}{c|}{8e-5} & 2e-3 \\ \hline
weight decay & \multicolumn{6}{c}{0} \\ \hline
batch size & \multicolumn{4}{c|}{64} & \multicolumn{1}{c|}{16} & 64 \\ \hline
optimizer & \multicolumn{6}{c}{AdamW~\cite{loshchilov2017decoupled}} \\ \hline
epochs & \multicolumn{4}{c|}{10} & \multicolumn{1}{c|}{6} & 5 \\ \hline
scheduler & \multicolumn{6}{c}{cosine} \\ \hline
warmup ratio & \multicolumn{5}{c|}{0.0075} & 0.03 \\ \hline
attention imp. & \multicolumn{6}{c}{Flash Attention 2~\cite{dao2023flashattention2}} \\ \hline
gradient ckpt. & \multicolumn{6}{c}{True} \\ \hline
ZeRO stage & \multicolumn{6}{c}{2} \\ \hline
dtype & \multicolumn{6}{c}{bfloat16} \\ \toprule[1.1pt]
\end{tabular}
}
\caption{Detailed training recipe, FWD: forward reaction prediction, REGT: reagent prediction, RTSTS: retrosynthesis, PRPT: property prediction, MCP: molcap, Pre.: Pretraining}
\label{tab:traindetail}
\end{table}
A detailed training configuration is listed in Table~\ref{tab:traindetail}, we conduct all of our experiments on Ubuntu-22.04, with CUDA 12.1 and PyTorch 2.4.1.

\section{Implementation Details}
\subsection{Graph Feature}
For graph features produced by our multi-level GNN, the token length of node and motif level feature varies with different molecule input, we use a Nested Tensor to store a zero padded tensor and it's corresponding mask. When using no reduction, we use mask to remove the padding tensor before the feature is fed into LLM, when average pooling is used, we use mask to calibrate the number of elements that is not padded.

\subsection{Chat Template}
The chat template of the 1B sized LLaMA-3.2 is as follows
\begin{tcolorbox}[notitle, rounded corners, colback=white, boxrule=1.5pt, boxsep=0pt, left=0.15cm, right=0.17cm, enhanced, shadow={2.5pt}{-2.5pt}{1.5pt}{opacity=5},toprule=2pt, before skip=0.65em, after skip=0.75em]
\noindent{\tt <|begin\_of\_text|><|start\_header\_id|>}system\\{\tt <|end\_header\_id|>}
\newline\newline
A chat between a curious user and an artificial intelligence assistant. The assistant gives helpful, detailed, and polite answers to the user's questions.{\tt <|eot\_id|>}

{\tt <|start\_header\_id|>}user{\tt <|end\_header\_id|>}
\newline\newline
{\tt <image>}$\backslash$n Instructions.{\tt <|eot\_id|>}

{\tt <|start\_header\_id|>}assistant{\tt <|end\_header\_id|>}
\newline\newline
Response.{\tt <|eot\_id|>}
\end{tcolorbox}

\section{Metrics}
\paragraph{Exact Match.}
Exact Match Score measures whether two SMILES strings strictly represent the same molecular structure. Specifically: Score of 1: The two SMILES are exactly the same after normalization, indicating that they represent the same molecule. Score of 0: The two SMILES are different after normalization, indicating that they represent different molecules.

\paragraph{Levenshtein Score.}
The Levenshtein Score measures the minimum number of edit operations required to convert one SMILES string into another. Typically, the edit operations include (1) Insertion: Inserting a character at a certain position. (2) Deletion: Deleting a character at a certain position. (3) Substitution: Replacing a character at a certain position with another character.

\paragraph{MACCS Similarity.}
MACCS Similarity is a method used in cheminformatics to evaluate and compare molecular structure similarity. It is based on MACCS keys, a predefined set of structural features developed by the Molecular ACCess System, which are used to describe and represent key substructures of molecules. The similarity between molecules is calculated by determining the presence or absence of these structural features.

\paragraph{RDK Similarity.}
RDK Similarity typically refers to the evaluation and calculation of molecular similarity using fingerprints generated by RDKit, an open-source cheminformatics toolkit.

\paragraph{Morgan Similarity.}
Morgan Similarity is a method used to assess and quantify structural similarity between molecules, based on Morgan fingerprints.

\end{document}